\documentclass[sn-mathphys-num]{sn-jnl}

\usepackage[utf8]{inputenc}
\usepackage[T1]{fontenc}

\usepackage{graphicx}
\usepackage{amsmath,amssymb,amsfonts}
\usepackage{amsthm}
\usepackage{mathtools}
\usepackage{mathrsfs}

\usepackage{booktabs}
\usepackage{multirow}
\usepackage{float}
\usepackage{stfloats}

\usepackage{subcaption}

\usepackage{algorithm}
\usepackage{algorithmic}

\usepackage{listings}

\usepackage[table]{xcolor}
\newcommand{\best}[1]{\cellcolor{gray!20}\textbf{#1}}
\usepackage{wrapfig}

\usepackage[capitalize,noabbrev]{cleveref}

\usepackage{geometry}

\theoremstyle{thmstyleone}%
\theoremstyle{thmstyletwo}%
\theoremstyle{thmstylethree}%

\begin{document}

\title[QE-Catalytic]{QE-Catalytic: A Graph–Language Multimodal Base Model for Relaxed-Energy Prediction in Catalytic Adsorption}


\author[1,2,4]{\fnm{Yanjie} \sur{Li}}\email{liyanjie@semi.ac.cn}
\author[2,3]{\fnm{Jian} \sur{Xu}}\email{jian.xu@ia.ac.cn}
\author[2,5]{\fnm{Xueqing} \sur{Chen}}\email{xqchen@cnic.cn}
\author[1]{\fnm{Lina} \sur{Yu}}\email{yulina@semi.ac.cn}
\author[2,3]{\fnm{Shiming} \sur{Xiang}}\email{smxiang@nlpr.ia.ac.cn}
\author*[1,2,4]{\fnm{Weijun} \sur{Li}}\email{wjli@semi.ac.cn}
\author*[2,3]{\fnm{Cheng-lin} \sur{Liu}}\email{liucl@nlpr.ia.ac.cn}

\affil[1]{\orgdiv{AnnLab}, \orgname{Institute of Semiconductors, Chinese Academy of Sciences}, \orgaddress{\city{Beijing}, \country{China}}}
\affil[2]{\orgdiv{Zhongguancun Academy}, \orgname{Zhongguancun Academy}, \orgaddress{\city{Beijing}, \country{China}}}
\affil[3]{\orgdiv{State Key Laboratory of Multimodal Artificial Intelligence Systems}, \orgname{Institute of Automation, Chinese Academy of Sciences}, \orgaddress{\city{Beijing}, \country{China}}}
\affil[4]{\orgdiv{School of Electronic, Electrical and Communication Engineering}, \orgname{University of Chinese Academy of Sciences}, \orgaddress{\city{Beijing}, \country{China}}}
\affil[5]{\orgdiv{Computer Network Information Center}, \orgname{Chinese Academy of Sciences}, \orgaddress{\city{Beijing}, \country{China}}}


\abstract{
Adsorption energy is a key descriptor of catalytic reactivity. It is fundamentally defined as the difference between the relaxed total energy of the adsorbate--surface system and that of an appropriate reference state; therefore, the accuracy of relaxed-energy prediction directly determines the reliability of machine-learning-driven catalyst screening. E(3)-equivariant graph neural networks (GNNs) can natively operate on three-dimensional atomic coordinates under periodic boundary conditions and have demonstrated strong performance on such tasks. In contrast, language-model-based approaches, while enabling human-readable textual descriptions and reducing reliance on explicit graph---thereby broadening applicability---remain insufficient in both adsorption-configuration energy prediction accuracy and in distinguishing ``the same system with different configurations,'' even with graph-assisted pretraining in the style of GAP-CATBERTa.

To this end, we propose QE-Catalytic, a multimodal framework that deeply couples a large language model (\textbf{Q}wen) with an E(3)-equivariant graph Transformer (\textbf{E}quiformer-V2), enabling unified support for adsorption-configuration property prediction and inverse design on complex catalytic surfaces. During prediction, QE-Catalytic jointly leverages three-dimensional structures and structured configuration text, and injects ``3D geometric information'' into the language channel via graph--text alignment, allowing it to function as a high-performance text-based predictor when precise coordinates are unavailable, while also autoregressively generating CIF files for target-energy-driven structure design and information completion. On OC20, QE-Catalytic reduces the MAE of relaxed adsorption energy from 0.713~eV to 0.486~eV, and consistently outperforms baseline models such as CatBERTa and GAP-CATBERTa across multiple evaluation protocols.
}

\keywords{Multimodal Large Language Models, Catalytic materials; Property Prediction, Relaxed Energy Prediction, Adsorption Energy Prediction}

\maketitle

\section{Introduction}
\label{Introduction}

In heterogeneous catalysis, adsorption energy is a key reactivity descriptor for catalyst screening. It essentially corresponds to the minimum-energy state attained by a given adsorbate--catalyst system across a diverse set of adsorption configurations that differ in adsorption sites, orientations, and related geometric degrees of freedom. Because the differences among these configurations are often subtle and their energies can be highly close---exhibiting near-degeneracy on the order of 0.1--0.3~eV around the global minimum---a model must not only predict the energy of an individual configuration, but also reliably discriminate near-degenerate configurations and identify the global minimum to ensure trustworthy adsorption-energy determination. However, achieving this objective within the DFT framework typically requires large-scale configuration enumeration and geometric relaxation, leading to prohibitive computational cost. Consequently, developing machine-learning models that can serve as efficient surrogates for DFT, accurately predicting relaxed energies across adsorption configurations and subsequently determining the minimum-energy state, has become a central direction for data-driven catalyst screening.

In recent years, E(3)-equivariant graph neural networks (graph neural networks, GNNs), such as the Equiformer family, have achieved remarkable performance in property prediction for atomistic systems. These models act directly on three-dimensional atomic coordinates under periodic boundary conditions and learn geometric information---including local coordination environments, bond lengths, and bond angles---without explicitly constructing hand-crafted features, reaching state-of-the-art performance for energy and force prediction on large-scale catalysis adsorption benchmarks such as OC20\cite{oc20}. However, this advantage also implies a strong dependence on high-quality three-dimensional structures: on the one hand, both training and inference require accurate or approximately relaxed atomic coordinates; on the other hand, it is difficult to naturally integrate extensive prior knowledge that exists in textual form---such as composition, Miller indices, and even experimental conditions---into the learned representations. In practical catalysis research, experiments and the literature typically provide ``symbolic descriptions'' (e.g., chemical formulas, facets, and adsorption-site types) rather than detailed three-dimensional coordinates, which limits the applicability of equivariant GNNs when high-quality geometric information is unavailable.

In contrast, Transformer- or BERT-based language models\cite{transsformer, bert} are inherently well-suited to processing human-readable string inputs. With appropriately designed text templates that encode information such as ``adsorbate + catalyst composition + Miller indices + configuration description'' into structured sequences, language models can learn statistical associations between adsorption energies and structure--property relationships directly in text space, without explicit three-dimensional coordinates or topology. Such approaches have shown initial promise in computational catalysis: they readily incorporate discrete information from the literature, databases, and experimental records, and provide a unified interface for diverse downstream tasks. However, because many geometrically distinct adsorption configurations can correspond to similar or even identical textual descriptions, text-only models struggle to precisely discriminate subtle differences in ``the same system with different configurations.'' Moreover, their energy-prediction accuracy remains limited.

To mitigate this issue, works such as GAP-CATBERTa introduced the idea of \emph{graph-assisted pretraining}: first extracting representations from three-dimensional atomic coordinates using mature graph neural networks, and then ``compressing'' them into the corresponding text features through self-supervised alignment, thereby indirectly injecting geometric information during training. This strategy improves, to some extent, the sensitivity of text models to configuration differences, enabling better adsorption-energy prediction performance under text-only inputs. Nevertheless, such methods still rely on a single text modality at inference: even when three-dimensional structural information is available, it cannot explicitly participate in prediction; furthermore, when different configurations share similar textual descriptions, the model remains unable to fully exploit fine-grained geometric differences to distinguish near-degenerate configurations.

The above observations expose a more fundamental issue: in complex catalytic systems, graph and text are not fully substitutable modalities. Three-dimensional atomic coordinates carry local structural information strictly constrained by E(3) symmetry, whereas textual descriptions are better suited to capturing high-level semantics such as composition, facets, defects, and processing conditions. If one relies on a single modality---either ``graph-only'' or ``text-only''---some information crucial for accurate adsorption-energy prediction is inevitably lost. This naturally raises the following question: can we build a truly multimodal model that simultaneously perceives three-dimensional graphs and structured text within a unified framework, maintains strong predictive performance even when the graph is missing, and can even perform inverse design by generating CIF files to complement missing information?

To address the above limitations, we propose QE-Catalytic: a multimodal framework that deeply couples an $E(3)$-equivariant graph Transformer with a large language model, enabling unified support for adsorption-configuration property prediction and inverse design on complex catalytic surfaces. Architecturally, QE-Catalytic adopts Equiformer-V2 as the geometric encoder to directly process atomistic structures under periodic boundary conditions, and uses structured configuration text---including the adsorbate, catalyst composition, and Miller indices, as well as primary/secondary neighbor configuration cues---as the language input. Through graph--text alignment training, the model constructs a shared graph--semantic latent space within a unified backbone, allowing geometric inductive bias to be injected into the language channel. Consequently, when three-dimensional coordinates are available, QE-Catalytic fully exploits geometric details; when precise coordinates are absent, it degrades to a high-performance text-based predictor. Moreover, leveraging the autoregressive generation capability of large language models, QE-Catalytic can automatically generate complete CIF files to complement missing configuration text information, and can even produce structural candidates such as CIFs conditioned on a target energy or a high-level system description, thereby forming an inverse structure-design pipeline oriented toward target adsorption energies. Overall, QE-Catalytic demonstrates a viable pathway for jointly leveraging three-dimensional graph and language representations in computational catalysis, providing methodological foundations for multimodal catalytic foundation models that achieve both high-accuracy property prediction and inverse-design capability.
The contributions of this paper are as follows:
\begin{itemize}
  \item \textbf{Multimodal architecture:} We propose a multimodal large language model architecture for catalytic materials property prediction, \textbf{QE-Catalytic}, which deeply couples an E(3)-equivariant graph Transformer (EquiformerV2) with a large language model. QE-Catalytic can process both 3D atomistic structures and structured configuration text within a unified framework, and supports text-only inference when geometric information is unavailable, as well as an inverse-design workflow based on CIF generation.
  \item \textbf{Improved loss-function design:} We propose a \emph{Max--Min tanh-gated multitask loss} (Max--Min Tanh-Gated Loss, MMTG-Loss) for jointly optimizing regression and classification/generation objectives. This loss is dominated by the larger sub-loss $L_{\max}$, and constructs a bounded gating factor from the smaller sub-loss $L_{\min}$ via $\tanh(L_{\min})$, thereby adaptively focusing on the ``hardest subtask'' during training and automatiadsorbate/surfacedient contributions from the regression loss and the cross-entropy loss.
  \item \textbf{Resolving the multi-configuration-within-the-same-system challenge:} To address the limited discriminative ability of prior language-model-based catalytic property predictors in scenarios involving ``the same adsorbate--catalyst system with different adsorption configurations,'' we substantially enhance the model's capability to distinguish configurations within the same system and to accurately predict subtle energy differences through graph--text alignment and multimodal joint training.
  \item \textbf{Multimodal data:} We curate and construct a multimodal dataset for catalytic materials property prediction, which consists of three-dimensional adsorption configurations under periodic boundary conditions and their corresponding structured textual descriptions organized using a three-part template (adsorbate / surface / configuration). In total, the dataset comprises 370,000 graph–language paired samples.
\end{itemize}

\section{Related Work}

\subsection{Energy prediction with graph neural networks}

Modeling three-dimensional atomistic systems as graphs and introducing equivariant inductive biases such as $E(n)$-, $SE(3)$-, and $E(3)$-equivariance in graph neural networks (GNNs) has led to substantial progress in catalysis and materials property prediction. Early representatives such as SchNet \cite{schutt2018schnet} modeled geometric information of molecules/materials via continuous-filter convolutions; subsequently, DimeNet/DimeNet++ \cite{klicpera2020dimenet,klicpera2020dimenetpp} introduced directional message passing and markedly improved the modeling of angle-dependent interactions. GemNet/GemNet-OC \cite{gasteiger2021gemnet,gasteiger2022gemnetoc} further incorporated multi-scale and higher-order geometric features, achieving strong performance on the OC20 suite of tasks. On equivariant message passing, methods such as SE(3)-Transformer \cite{fuchs2020se3transformer}, EGNN \cite{satorras2021egnn}, PaiNN \cite{schutt2021painn}, NequIP \cite{batzner2022nequip}, MACE \cite{batatia2022mace}, and Allegro \cite{musaelian2023allegro} have systematically introduced rotation-equivariant representations into molecular force fields and materials property prediction.

Building on this line, equivariant-Transformer-based Equiformer \cite{liao2023equiformer} and EquiformerV2 \cite{liao2024equiformerv2} combine irreducible representations (irreps) with graph attention. Equiformer achieves strong competitiveness on datasets such as QM9, MD17, and OC20. EquiformerV2, through design choices including the efficient tensor product from eSCN \cite{passaro2023escn}, separable $S^2$ activations, and separable layer normalization, significantly reduces computational cost under higher-order representations, and attains improved energy/force prediction accuracy and data efficiency on OC20/OC22. In addition, SCN \cite{zitnick2022scn}, SEGNN \cite{brandstetter2022segnn}, and the Transformer-based Graphormer \cite{ying2021graphormer} have also reported strong results on related tasks.

Overall, purely GNN/equivariant approaches have a natural advantage in \emph{capturing fine-grained three-dimensional geometric details}, but they typically \emph{struggle to directly leverage textual information} (e.g., experimental conditions and qualitative descriptions from the literature), which limits their ability to incorporate language modalities to supplement cross-system knowledge.

\subsection{Energy prediction with language models (text-driven)}

To overcome the limited utilization of textual knowledge in purely structure-based paradigms, researchers have begun to explore \emph{text-centric} property prediction. For example, MOFormer \cite{moformer2023} encodes MOFs as strings (MOFid); TransPolymer \cite{transpolymer2023} takes SMILES and polymer attributes as inputs; composition-driven Roost \cite{goodall2020roost} and CrabNet \cite{wang2021crabnet} achieve favorable generalization using only chemical formulas/compositions. In addition, materials-domain language models (e.g., MatSciBERT/MatBERT) have been used to extract knowledge from the literature and support downstream tasks \cite{gupta2022matscibert,matbert2021}.

In the context of catalytic adsorption energies, CatBERTa regresses adsorption-configuration energies from human-readable descriptions of catalytic systems \cite{ock2023catalyst}, avoiding a strong dependence on precise coordinates. However, because \emph{the same textual description often corresponds to multiple adsorption configurations with near-identical energies}, text-only models have limited capacity to resolve \emph{subtle geometric differences}, thereby constraining their base accuracy. GAP-CatBERTa aligns the structural-embedding knowledge of EquiformerV2 into the textual embedding space through \emph{graph-assisted contrastive pretraining}, and subsequently achieves lower MAE and higher $R^2$ in \emph{text-only} downstream fine-tuning \cite{ock2024gapcatberta}. Nevertheless, such methods still operate primarily in a single text modality at \emph{inference} time and do not explicitly incorporate real 3D configurations, leaving an ambiguity challenge when ``multiple structures correspond to the same text.''
\subsection{Generation and inverse design with large language models in materials science}
Early methods largely relied on graph-based or equivariant-network generative frameworks, such as G-SchNet \cite{g-schnet}, CDVAE \cite{cdvae}, DiffCSP \cite{diffcsp}, MatterGen \cite{mattergen}, and MatterSim \cite{mattersim}. These approaches learn the distribution of crystalline structures in three-dimensional space and leverage energy or property predictors to enable target-guided structural search.
Large language models (LLMs) exhibit \emph{conditional generation/inverse-design} potential under the paradigm that ``structure is text.'' CrystaLLM \cite{antunes2024crystallm} treats CIF files as a training corpus for autoregressive learning, and couples the model with a property predictor or search strategy to generate crystal candidates that satisfy target properties, demonstrating the ability to \emph{infer structures from target properties}. Beyond LLMs, crystal generation based on diffusion and flows has also advanced rapidly; works such as DiffCSP and CDVAE improve the sampling and recovery of feasible structures from the perspectives of symmetry and energy distributions \cite{xie2021crystalnet,zhou2022diffcsp,jiao2021cdvae}. These methods provide a toolchain for \emph{structural hypothesis generation} in ``structure-missing'' scenarios: candidate structures (e.g., CIF) are first generated, then converted into inputs compatible with text/graph encoders for property evaluation, thereby closing the loop of an inverse-design workflow of ``generate--evaluate--screen.''

\paragraph{Summary}
In summary, graph-based methods excel at modeling three-dimensional graph but struggle to integrate language knowledge; language-based methods readily absorb semantics and experimental context but are insensitive to critical geometric differences; and generative LLMs and diffusion models provide new avenues for ``structure-missing'' inputs and inverse design. 
Our approach adopts EquiformerV2 as the 3D feature extractor and Qwen2.5-VL as the language backbone, and jointly models \emph{three-dimensional atomistic structures} and \emph{textual descriptions of catalytic systems} within a \emph{unified multimodal framework}: we first freeze the LLM for cross-modal alignment and then perform full-parameter fine-tuning to achieve accurate prediction of relaxed energy. Compared with existing methods, our work offers: (1) \textbf{joint text + 3D-structure} prediction, combining geometric discrimination with semantic completion; (2) in \textbf{structure-missing} scenarios, the model can naturally degrade to a ``text-only'' mode and achieves better performance than CatBERTa trained purely on text; and (3) \textbf{inverse-design} potential: when catalytic-system information is incomplete, the model can generate CIF files from known information to complement missing inputs, enabling more accurate property prediction.

\begin{table*}[t]
  \centering
  \caption{Performance comparison between QE-Catalytic and text-based baseline models on OC20 and OC20-Dense. QE-Catalytic* denotes a model trained multimodally but using only text inputs during inference.}
  \label{tab-1}
  \resizebox{\linewidth}{!}{%
  \begin{tabular}{l l l cc cc}
    \toprule
    & \multicolumn{1}{c}{Pretrain Data}
    & \multicolumn{1}{c}{Fine-tuning Data}
    & \multicolumn{2}{c}{Prediction Results}
    & \multicolumn{2}{c}{Improvement from CatBERTa} \\
    \cmidrule(lr){4-5} \cmidrule(lr){6-7}
    &  &  & MAE [eV] (\(\downarrow\)) & \(R^2\) [-] (\(\uparrow\))
       & MAE (\%) (\(\downarrow\)) & \(R^2\) (\%) (\(\uparrow\)) \\
    \midrule
    \multirow{2}{*}{CatBERTa}
      & --           & OC20 (460k)          & \(0.713 \pm 0.014\) & \(0.584 \pm 0.014\) & --     & --    \\
      & --           & OC20-Dense (16k)     & \(0.542 \pm 0.011\) & \(0.712 \pm 0.008\) & --     & --    \\
    \midrule
    \multirow{2}{*}{GAP-CatBERTa}
      & OC20 (460k)  & OC20 (460k)          & \(0.643 \pm 0.020\) & \(0.691 \pm 0.015\) & \(-9.82\) & \(+18.32\) \\
      & OC20 (460k)  & OC20-Dense (16k)     & \(0.502 \pm 0.010\) & \(0.764 \pm 0.008\) & \(-7.38\) & \(+7.30\)  \\
    \midrule
    \multirow{2}{*}{QE-Catalytic*}
      & OC20 (460k)  & OC20 (460k)          & \(0.584 \pm 0.016\) & \(0.724 \pm 0.017\) & \(-18.1\) & \(+24.0\) \\
      & OC20 (460k)  & OC20-Dense (16k)     & \(0.482 \pm 0.011\) & \(0.792 \pm 0.010\) & \(-12.0\) & \(+11.2\)  \\
    \midrule
    \multirow{2}{*}{QE-Catalytic}
      & OC20 (460k)  & OC20 (460k)
      & \best{$0.486 \pm 0.018$} & \best{$0.788 \pm 0.012$}
      & \best{$-31.8$} & \best{$+35.0$} \\
      & OC20 (460k)  & OC20-Dense (16k)
      & \best{$0.427 \pm 0.014$} & \best{$0.818 \pm 0.012$}
      & \best{$-21.2$} & \best{$+14.9$}  \\
    \bottomrule
  \end{tabular}%
  }
\end{table*}
\section{Results}
In this paper, we evaluate QE-Catalytic from multiple perspectives. In Section~5.1, we compare QE-Catalytic with recent language-model-based baselines. In Section~5.2, we evaluate QE-Catalytic against traditional graph-based baselines. In Section~5.3, we visualize the feature-alignment status in the latent space and the model's discriminability for samples from the same system with different configurations. In Section~5.4, we test whether prediction accuracy can be improved in scenarios where crystal-structure information is missing by first generating a CIF with the model and then complementing the input information. In Section~5.5, we conduct ablation studies on different output heads.

\subsection{Comparison with language-model-based baselines}
We first compare QE-Catalytic with recent language-model-based approaches for adsorption-energy prediction, including the original CatBERTa and GAP-CatBERTa, which incorporates graph-assisted pretraining (GAP). To ensure comparability, under the same pretraining/fine-tuning data setting, we report the MAE and $R^2$ metrics of each method on OC20 and OC20-Dense; the results are shown in \cref{tab-1}.

CatBERTa achieves an MAE of $0.713\pm0.014$\,eV and $R^2=0.584\pm0.014$ on OC20, and an MAE of $0.542\pm0.011$\,eV and $R^2=0.712\pm0.008$ on OC20-Dense. With GAP, GAP-CatBERTa shows clear improvements on both datasets; for example, on OC20 it reduces the MAE to $0.643\pm0.020$\,eV and increases $R^2$ to $0.691\pm0.015$, corresponding to a $9.82\%$ relative reduction in MAE and an $18.32\%$ relative improvement in $R^2$. These results indicate that ``compressing'' three-dimensional geometric information into textual features during pretraining indeed enhances the text model's sensitivity to configuration differences.

Under the same training data setting on OC20 (460k), both QE-Catalytic* (multimodally trained but text-only at inference) and QE-Catalytic (using 3D+text at inference) substantially outperform the above text baselines on both benchmarks. Taking OC20 as an example, QE-Catalytic* achieves an MAE of $0.584\pm0.016$\,eV and $R^2=0.724\pm0.017$, further reducing MAE by roughly $10\%$ relative to GAP-CatBERTa and improving $R^2$ by about 4 percentage points. QE-Catalytic further reduces MAE to $0.486\pm0.018$\,eV and increases $R^2$ to $0.788\pm0.012$, corresponding to a $31.8\%$ MAE reduction and a $35.0\%$ $R^2$ improvement over the original CatBERTa. Results on OC20-Dense exhibit the same trend: QE-Catalytic reduces MAE from $0.542$\,eV to $0.427\pm0.014$\,eV and improves $R^2$ from $0.712$ to $0.818\pm0.012$, yielding relative improvements of $21.2\%$ and $14.9\%$, respectively.
Overall, QE-Catalytic achieves clear advantages over other language-model-based approaches. On the one hand, multimodal training enables QE-Catalytic* to outperform GAP-CatBERTa even under text-only inference, demonstrating that geometric features have been effectively injected into the language channel. To further verify that the textual and geometric features are indeed highly aligned in the feature space (i.e., geometric features are effectively injected into the language channel), we visualize the cross-correlation matrix for a subset of paired samples, as shown in Fig.~\ref{hamap}. On the other hand, when graph and text are jointly used for inference, QE-Catalytic further improves prediction accuracy substantially.
\subsection{Comparison with classical baselines}
We further compare QE-Catalytic with a suite of classical atomistic graph neural network baselines, including SchNet, PaiNN, DimeNet++, GemNet-OC, as well as the equivariant-Transformer-based Equiformer and EquiformerV2. All models are trained on OC20 (460k) and evaluated on the same test split for relaxed adsorption-energy prediction; the results are reported in \cref{tab-2}.
\begin{wraptable}{r}{0.49\textwidth}
  \centering
  \vspace{-10pt}
  \caption{Performance comparison between QE-Catalytic and classical GNN baselines on OC20 (460k).}
  \label{tab-2}
  \resizebox{\linewidth}{!}{%
  \begin{tabular}{l c cc}
    \toprule
    & \multicolumn{1}{c}{Training Data}
    & \multicolumn{2}{c}{Prediction Results} \\
    \cmidrule(lr){3-4}
    &  & MAE [eV] (\(\downarrow\)) & \(R^2\) [-] (\(\uparrow\)) \\
    \midrule
    GemNet-OC    & OC20 (460k)          & \(0.841 \pm 0.015\) & \(0.428 \pm 0.014\) \\
    SchNet       & OC20 (460k)          & \(0.988 \pm 0.012\) & \(0.324 \pm 0.014\) \\
    PaiNN        & OC20 (460k)          & \(0.926 \pm 0.021\) & \(0.338 \pm 0.015\) \\
    DimeNet++    & OC20 (460k)          & \(0.713 \pm 0.011\) & \(0.584 \pm 0.014\) \\
    Equiformer   & OC20 (460k)          & \(0.814 \pm 0.010\) & \(0.439 \pm 0.015\) \\
    EquiformerV2 & OC20 (460k)          & \(0.682 \pm 0.006\) & \(0.651 \pm 0.015\) \\
    QE-Catalytic*     & OC20 (460k)          & \(0.584 \pm 0.016\) & \(0.724 \pm 0.017\) \\
    QE-Catalytic      & OC20 (460k)
                 & \best{$0.486 \pm 0.018$} & \best{$0.788 \pm 0.012$} \\
    \bottomrule
  \end{tabular}%
  }
\end{wraptable}
As shown in \cref{tab-2}, among these conventional GNNs, EquiformerV2 achieves the best performance, outperforming earlier methods such as SchNet, PaiNN, and GemNet-OC. DimeNet++ attains an MAE of $0.713\pm0.011$\,eV and $R^2=0.584\pm0.014$, which is numerically comparable to CatBERTa. Without changing the training data, QE-Catalytic reduces the MAE to $0.486\pm0.018$\,eV and improves $R^2$ to $0.788\pm0.012$, indicating that introducing multimodal modeling and alignment yields substantial gains while retaining equivariant geometric perception. Moreover, since the geometric branch of QE-Catalytic is itself based on EquiformerV2, we attribute these improvements primarily to the following: multimodal training enables geometric information to be more thoroughly integrated into the semantic representations of the LLM backbone, thereby better capturing subtle local-environment differences and suppressing prediction noise in dense configuration spaces.
In summary, QE-Catalytic not only consistently outperforms existing language-model baselines under the ``text+3D'' paradigm, validating the effectiveness of deeply embedding an equivariant geometric encoder into a multimodal large language model, but also exhibits strong advantages in the ``text-only'' setting. In particular, it significantly surpasses the current equivariant GNNs.

\subsection{Comparative experiments for the Max--Min tanh-gated multitask loss}
\label{subsec:loss_ablation}
\begin{wrapfigure}{r}{0.49\textwidth}
  \centering
  \vspace{-18pt}
  \includegraphics[width=\linewidth]{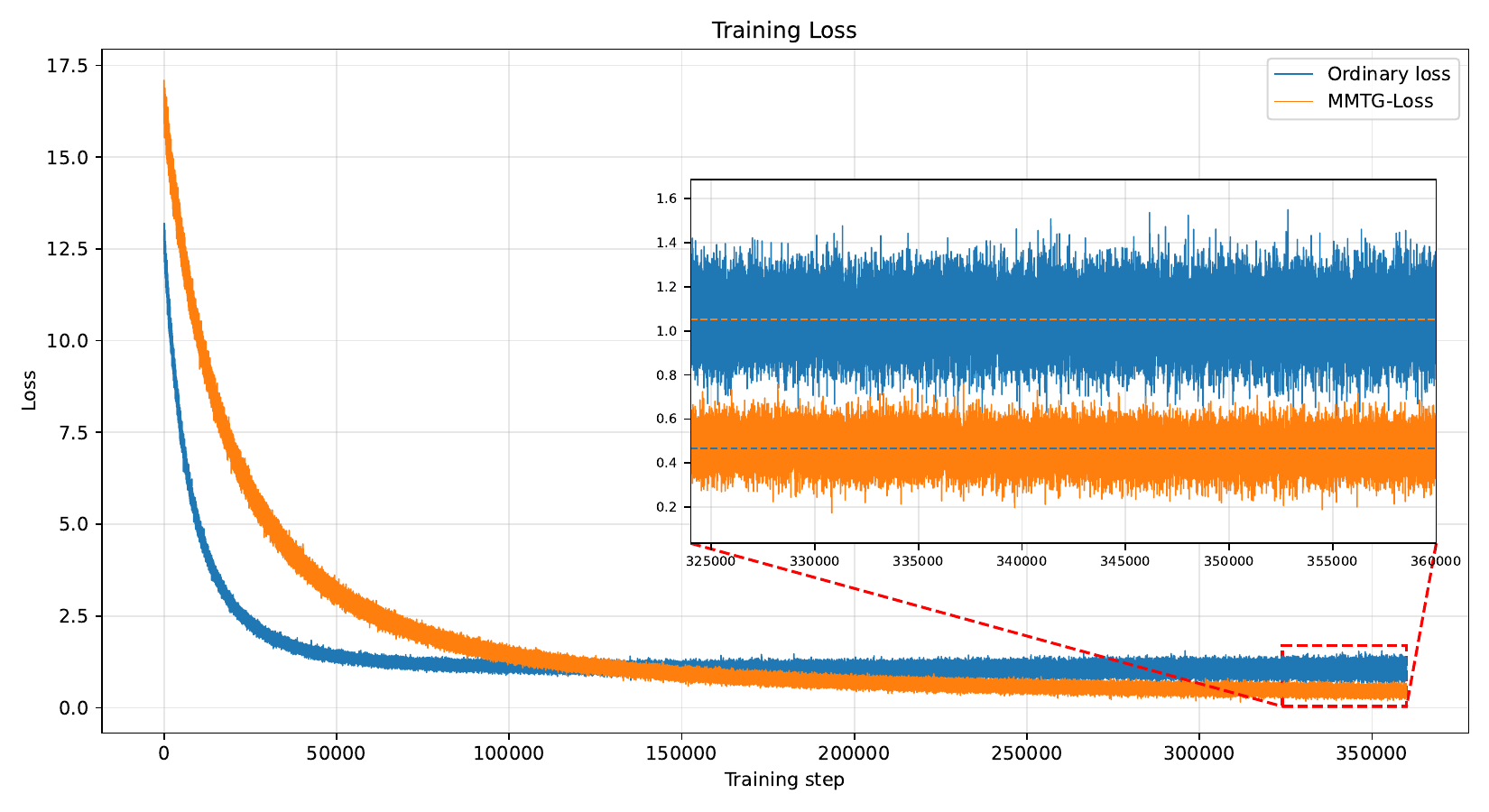}
  \caption{Comparison of training dynamics between the ordinary loss and MMTG-Loss. The ordinary loss decreases faster in the early stage but quickly enters a relatively high plateau. In contrast, MMTG-Loss exhibits a smoother overall descent, continues to decrease in later stages, and converges to a lower final loss level.}
  \label{fig-loss}
\end{wrapfigure}
To systematically evaluate the advantages of the \emph{Max--Min tanh-gated multitask loss} (Max--Min Tanh-Gated Loss, hereafter abbreviated as MMTG-Loss) over the conventional weighted-sum loss, we first conduct a controlled comparison in a simulated training setting. Specifically, we train the same model on the same data using two different loss formulations, and the loss-decrease curves are shown in Fig.~\ref{fig-loss}. From the overall convergence trends, the conventional weighted-sum loss decreases faster in the early stage, but quickly enters a ``plateau'' after several epochs, where the loss value becomes difficult to further reduce, exhibiting a typical pattern of ``fast start, early convergence, but a relatively high plateau.'' In contrast, although MMTG-Loss decreases slightly more slowly at the beginning, it maintains a persistent and stable downward trend throughout training. The zoomed-in view embedded in the figure further shows that in the late stage (the last $\sim 10\%$ of steps), the curve corresponding to MMTG-Loss not only achieves a lower mean loss but also exhibits smaller fluctuations, indicating better long-horizon convergence and stability.

This observation is consistent with the design motivation of MMTG-Loss. By introducing
\(
L_{\max}, L_{\min}
\)
and using the bounded $\tanh(L_{\min})$ as a gating factor, the model always prioritizes the subtask with the larger current error (the dominant term $L_{\max}$), while allowing the smaller-error subtask to gently modulate the overall loss through the gating term. This avoids the situation in which a single subtask converges too quickly in the early stage and prematurely dominates the gradient signal. Compared with the conventional linearly weighted multitask loss, MMTG-Loss can adaptively balance the optimization progress of regression ($L_{\mathrm{MAE}}$) and classification/generation ($L_{\mathrm{CE}}$) objectives without requiring delicate hyperparameter tuning, leading to smoother training dynamics, a lower final composite loss, and improved downstream performance in practical tasks.

\subsection{Latent space and separability of multiple configurations within the same system}
\begin{figure*}[htp]
\centering
\includegraphics[width=136mm]{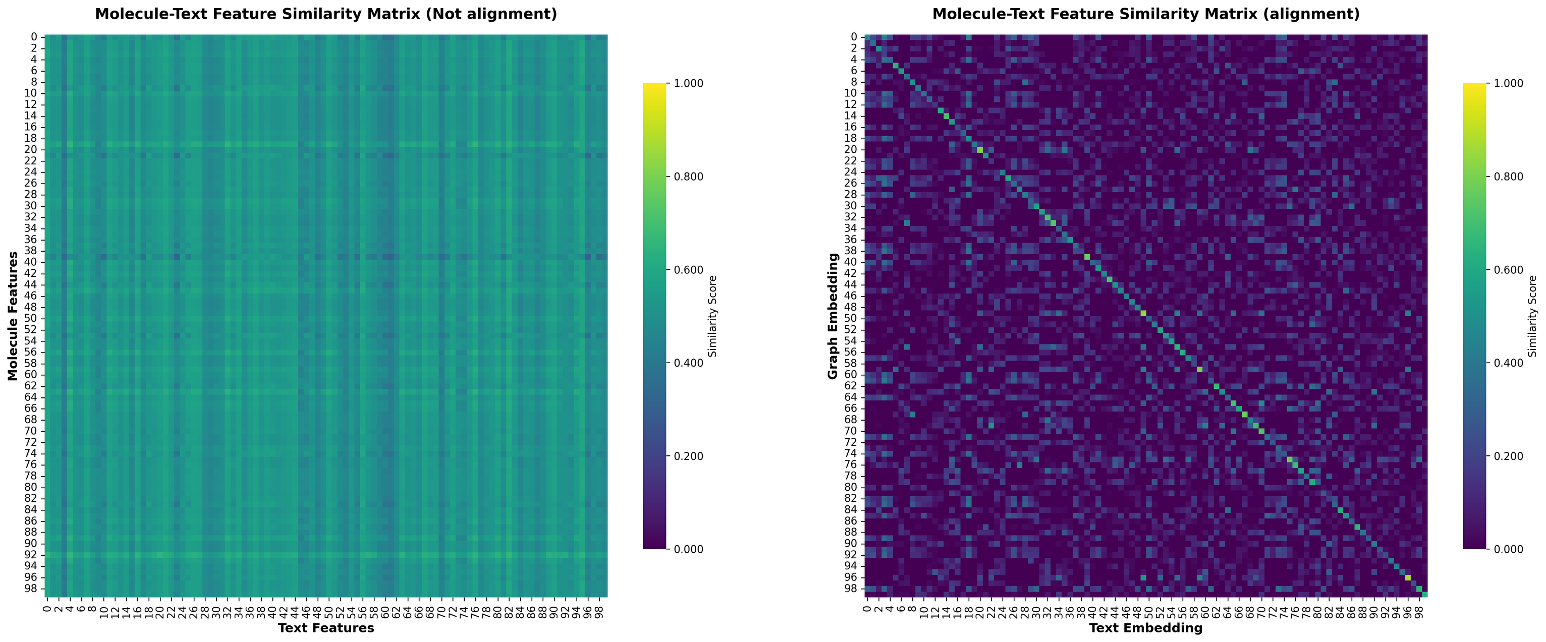}
\caption{
Illustration of cross-modal alignment. The figure shows a similarity matrix between geometric embeddings and text embeddings, with the horizontal and vertical axes corresponding to sample indices. After multimodal alignment, a clear diagonal structure emerges in the similarity matrix, indicating that the geometric and textual representations of the same configuration are highly consistent in the latent space, whereas similarities between different configurations are substantially lower.
}
\label{hamap}
\end{figure*}

\begin{figure*}[htp]
\centering
\includegraphics[width=134mm]{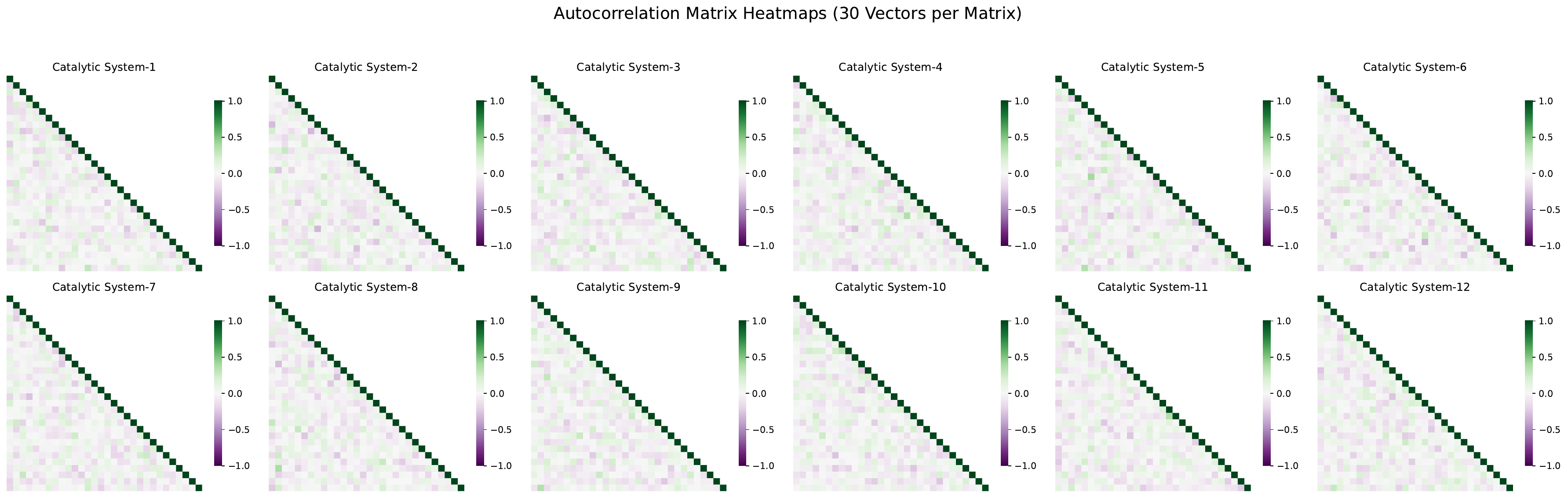}
\caption{
An example of autocorrelation similarity heatmaps for multi-configuration samples under the same catalytic system. Each subfigure corresponds to a fixed adsorbate--catalyst pair; the horizontal and vertical axes index different configurations, and the color denotes cosine similarity of QE-Catalytic latent embeddings. It can be observed that even within the same system, the representation similarity between different configurations is substantially separated, indicating that the model can discriminate local geometric differences rather than collapsing ``all configurations of the same system'' into nearly identical representations.
}
\label{fig_12}
\end{figure*}
During pretraining, we align the 3D molecular features produced by EquiformerV2 with the corresponding textual string features via contrastive learning. After multimodal training, even under text-only inputs, the model can perform inference within this shared latent space. Therefore, it is necessary to inspect the similarity between geometric and textual embeddings in the latent space to validate the alignment quality, and to further analyze the model's separability in the ``same-system multi-configuration'' scenario.

First, we visualize the similarity matrix between graph features and text features (see \cref{hamap}). The rows and columns correspond to geometric and textual embeddings, respectively, and the color denotes cosine similarity. As shown in the figure, after multimodal training a clear diagonal structure emerges in the similarity matrix, indicating that the model has successfully aligned the graph and text representations corresponding to the same adsorbate--catalyst configuration in the latent space, while similarities between different samples are significantly lower.
Second, to evaluate QE-Catalytic's discriminability under the condition of ``the same adsorbate--catalyst system with different configurations,'' we select 12 catalytic systems from the dataset, each containing multiple adsorption configurations, with the smallest system containing 31 configurations. To visualize representational differences between configurations, we randomly sample 30 configurations from each system, compute the autocorrelation similarity matrix among their latent embeddings, and plot the corresponding heatmaps, as shown in \cref{fig_12}.

From these heatmaps, the correlations among QE-Catalytic embeddings for different configurations within the same system neither degenerate into uniformly high similarity (i.e., ``all configurations look the same''), nor scatter in a completely unstructured manner. Instead, they exhibit a structured distribution that varies smoothly with graph/configuration changes. This suggests that QE-Catalytic can effectively separate representations of different configurations while preserving system-level semantic consistency, providing a representational basis for accurately distinguishing near-degenerate configurations.

\subsection{Energy prediction for unknown structures and the PIR metric}
An important advantage of language models and text-based representations is that they can, to some extent, bypass dependence on explicit atomistic structures. QE-Catalytic can handle textual descriptions with or without neighbor information; however, to achieve the best performance, the input strings typically need to include neighboring-atom information. 
To obtain better prediction performance in structure-missing scenarios, we explore generating the required configuration text using the same multimodal large language model based only on adsorbate and catalyst information. The core idea is to \emph{derive the final ``configuration'' segment from the first two segments (adsorbate and catalyst information)}.
Specifically, we fine-tune the LLM using CIF files corresponding to relaxed structures in the OC20 training data (the procedure is shown in \cref{fig-2}), so that given the first two text segments ``adsorbate + catalyst composition + Miller indices,'' it can autoregressively generate an indicative CIF file. To simplify generation, we terminate decoding when two consecutive newline characters are produced, and in post-processing we retain only samples whose elemental composition matches the expectation. The fine-tuned model takes inputs such as ``data CCH\textsubscript{3}</s>Al\textsubscript{12}As\textsubscript{12} (1 1 1)'' and outputs the corresponding CIF file containing indicative structural information.
Because the atomic coordinates in these generated CIFs may not be physically accurate, we do not use them directly as inputs to the GNN. Instead, following the procedure described above, we extract the local neighborhood from the generated CIF and convert it back into a three-part configuration string. In this way, when the configuration is unknown, QE-Catalytic can still be provided with an ``LLM-derived configuration string'' for energy prediction or candidate screening.
\begin{figure*}[htp]
    \centering
    \setlength{\belowcaptionskip}{-0.0cm}
       \subfloat[]{
       \includegraphics[width=0.625\linewidth]{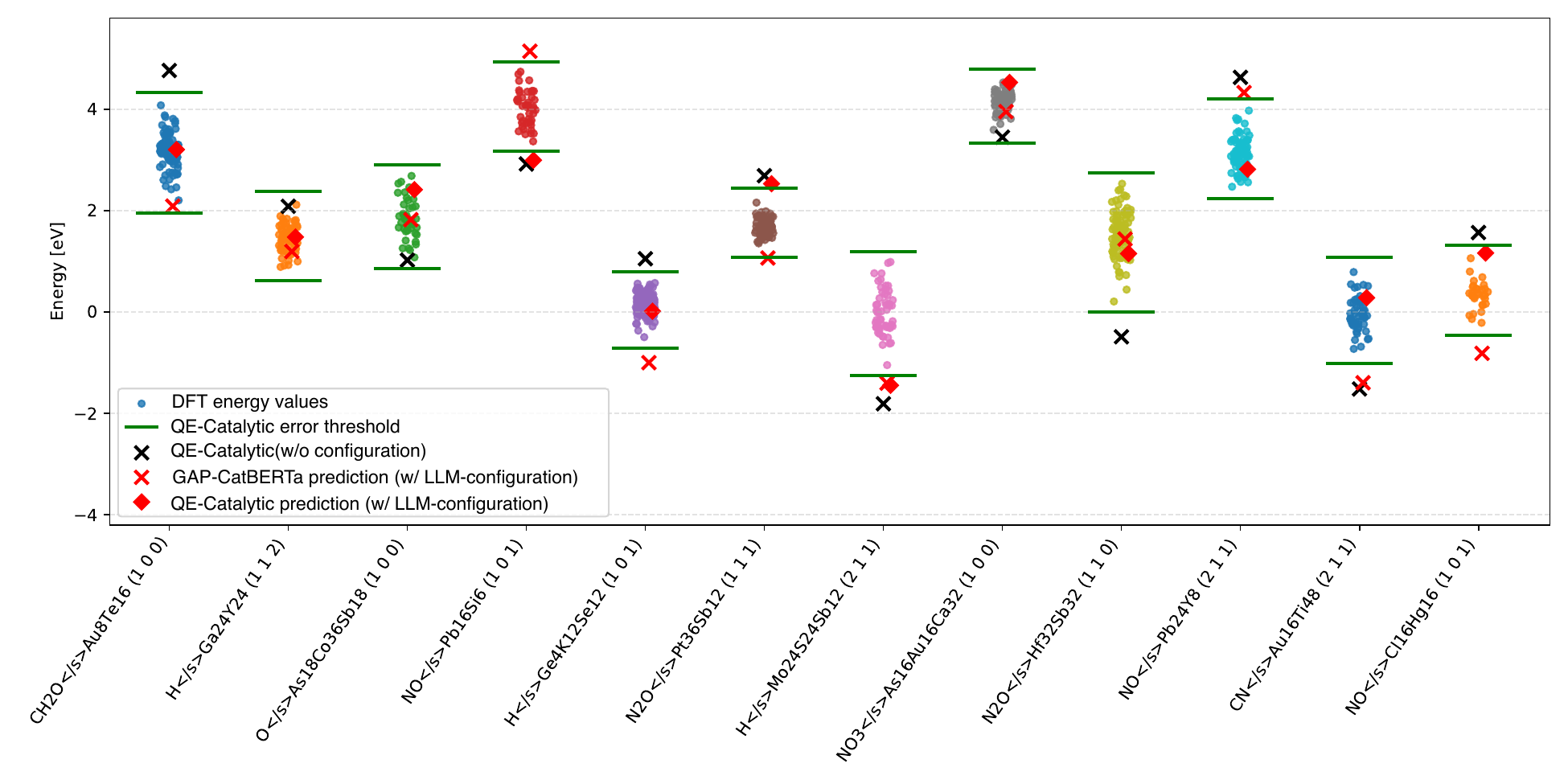} \label{PIRa}}
        \subfloat[]{
        \includegraphics[width=0.36\linewidth]{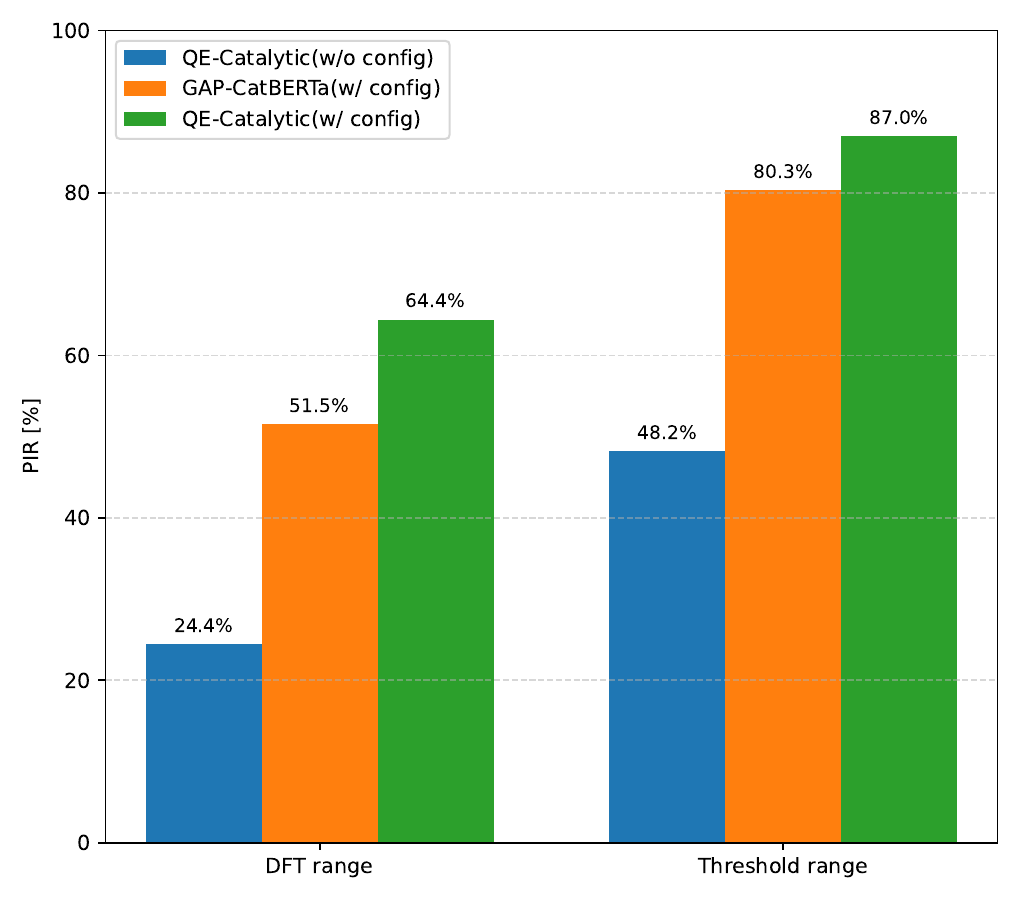}\label{PIRb}}
	  \caption{
   Performance gains from using LLM-derived configuration strings as inputs to QE-Catalytic. (a) Twelve representative examples selected from 66 adsorbate--catalyst pairs; colored dots indicate energies of different adsorption configurations under each pair. (b) Prediction Inclusion Ratio (PIR) over the 66 pairs, which quantifies the improvement in prediction accuracy after adding LLM-generated configuration strings (``config.'') to the input.
  }
\label{fig3}
\end{figure*}
To quantify the improvement brought by the LLM-derived configuration strings, we start from the OC20-Dense training set, which contains 235 unique adsorbate--catalyst pairs, and downsample 66 adsorbate--catalyst pairs according to the types and counts of elements in the adsorbate and the catalyst. The selected adsorbate and catalyst information is used as the initial prompt and fed into QE-Catalytic to generate the corresponding CIF (see Fig.~\ref{fig-2}). We run CIF generation multiple times for each pair and filter samples that satisfy a composition threshold, then convert them into three-part strings for energy prediction. Across these 66 pairs, OC20-Dense contains a total of 5,141 possible adsorption configurations. We use the Prediction Inclusion Ratio (PIR) as the evaluation metric:
\begin{equation}
\label{eq_pir}
PIR[\%] = \frac{N_{in\text{-}range}}{N_{total}} \times 100,
\end{equation}
where $N_{in\text{-}range}$ denotes the number of predictions that fall within the target energy range (accounting for CatBERTa's own error tolerance), and $N_{total}$ is the total number of predictions.

We compare PIR under two settings: using only the first two segments (without configuration) versus adding the LLM-generated configuration. As can be seen, across the 66 test pairs, introducing the LLM-derived configuration strings nearly doubles the probability that predictions fall within the target energy range. This indicates that even when the generated CIF structures are only indicative, the local-neighborhood information they provide is sufficient to substantially improve QE-Catalytic's prediction quality for relaxed adsorption energies.
\subsection{Ablation study on three training variants and the dual-head output}
\begin{wrapfigure}{r}{0.49\textwidth}
  \centering
  \vspace{-20pt}
  \includegraphics[width=\linewidth]{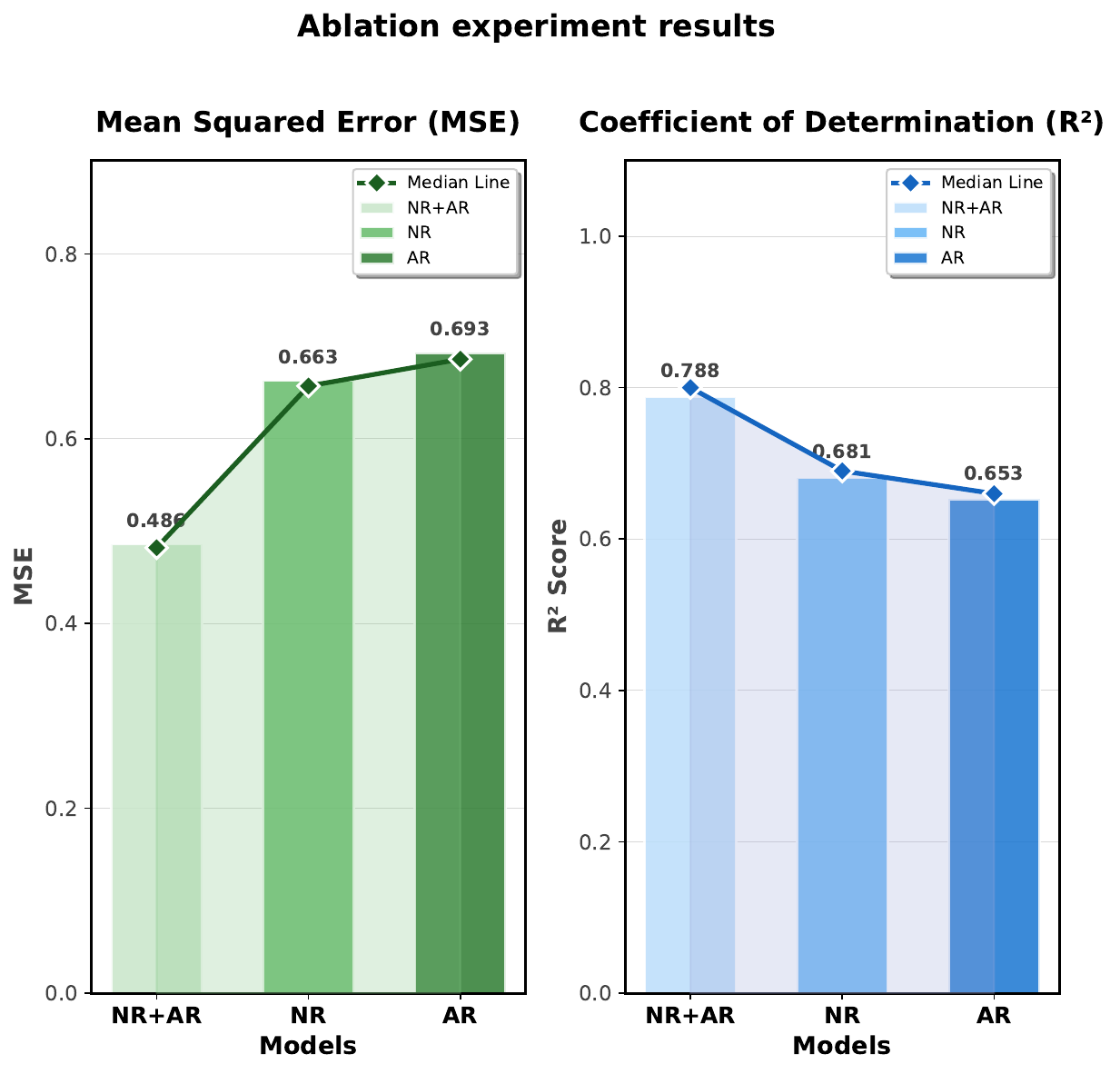}
  \caption{Ablation results for different output-head designs. The horizontal axis enumerates model variants (regression-only head, autoregressive-only head, and dual-head), and the vertical axis reports MAE and $R^2$. The dual-head design achieves slightly better performance on both metrics than either single-head variant, indicating that combining numerical regression with autoregressive generation benefits overall prediction performance.}
  \label{fig_ablation}
\end{wrapfigure}
The final QE-Catalytic output layer adopts a dual-head design: a numerical regression head and an autoregressive head. Both operate on the shared multimodal latent representation, performing regression and generation, respectively; the final prediction is taken as the average of the two outputs. To test the effectiveness of this design, we conduct ablation experiments with the following three training variants\cite{ab1} (results are shown in \cref{fig_ablation}):
\begin{enumerate}
    \item \textbf{Regression head only:} We remove the autoregressive head and retain only the numerical regression head to directly regress energies. This setting corresponds to the classical ``encoder--regressor'' architecture.
    \item \textbf{Autoregressive head only:} We remove the numerical regression head and retain only the autoregressive head, predicting energies by generating energy-related tokens and parsing them into numerical values.
    \item \textbf{Dual-head output (ours):} We keep both the numerical regression head and the autoregressive head, jointly optimize the two losses during training, and average the two head outputs as the final prediction during inference.
\end{enumerate}

Experimental results show that on OC20 and OC20-Dense, the dual-head design achieves slightly better MAE and $R^2$ than either single-head variant. Concretely, the regression-only model converges faster and is more stable, but exhibits limited extrapolation ability on certain complex systems. The autoregressive-only model is more strongly constrained by language priors during generation and tends to be slightly conservative when predicting extreme energy values. By simultaneously leveraging continuous regression signals and discrete generation signals, the dual-head structure provides complementary supervision and a regularization effect to some extent, resulting in improved overall performance.

\section{Conclusion and Discussion}

This paper presents QE-Catalytic, a multimodal framework that deeply couples an $E(3)$-equivariant graph Transformer with a large language model for adsorption-configuration energy prediction and inverse design on complex catalytic surfaces. By adopting EquiformerV2 as the geometric encoder and Qwen2.5-VL as the language backbone, and by constructing a shared graph--semantic latent space between them, QE-Catalytic can jointly process three-dimensional atomistic structures and structured textual descriptions within a single unified model. We design a three-stage training pipeline: we first freeze the LLM and perform graph--text alignment via contrastive learning; we then unfreeze the full model for multimodal joint pretraining; finally, we freeze the geometric branch and conduct instruction fine-tuning on the LLM. Experimental results on standard benchmarks such as OC20 and OC20-Dense demonstrate that QE-Catalytic substantially outperforms text-based models including CatBERTa and GAP-CatBERTa under both text-only and multimodal settings, and also surpasses equivariant GNN baselines such as SchNet and EquiformerV2.

Methodologically, QE-Catalytic exemplifies a multimodal scientific modeling paradigm of ``graph as perception, language as interface'': the geometric encoder captures local coordination and continuous structural details in an equivariant manner, while the language model integrates textual system information, experimental conditions, and knowledge priors, and exposes prediction and generation capabilities through an instruction-following interface. Multimodal alignment injects geometric information into the language channel, enabling strong performance under text-only inference when graph is missing; conversely, the generative capability of the language channel provides a practical tool for structural hypothesis generation and inverse design. In structure-unknown scenarios, QE-Catalytic can first generate an indicative CIF via the LLM, then extract a local configuration string for energy prediction, substantially increasing the probability that predictions fall within a target energy range.

Future directions include extending QE-Catalytic to more complex reaction networks and multi-step reaction pathways, and integrating energy prediction with kinetics modeling and force-field learning; exploring multitask learning within a unified multimodal framework for energy, forces, uncertainty estimation, and adsorption-site classification; further incorporating experimental data and literature-derived information to improve transferability and reliability under realistic operating conditions; leveraging reinforcement-learning-based fine-tuning techniques such as DPO\cite{dpo} and GRPO\cite{grpo} to improve the quality of CIF generation and enable genuinely effective inverse design; and validating the scalability of QE-Catalytic as a ``scientific multimodal foundation model'' on larger-scale materials systems.
\section{Method}

\subsection{Overall framework}
\begin{figure*}[htp]
\centering
\hspace{-8pt}
\includegraphics[width=134mm]{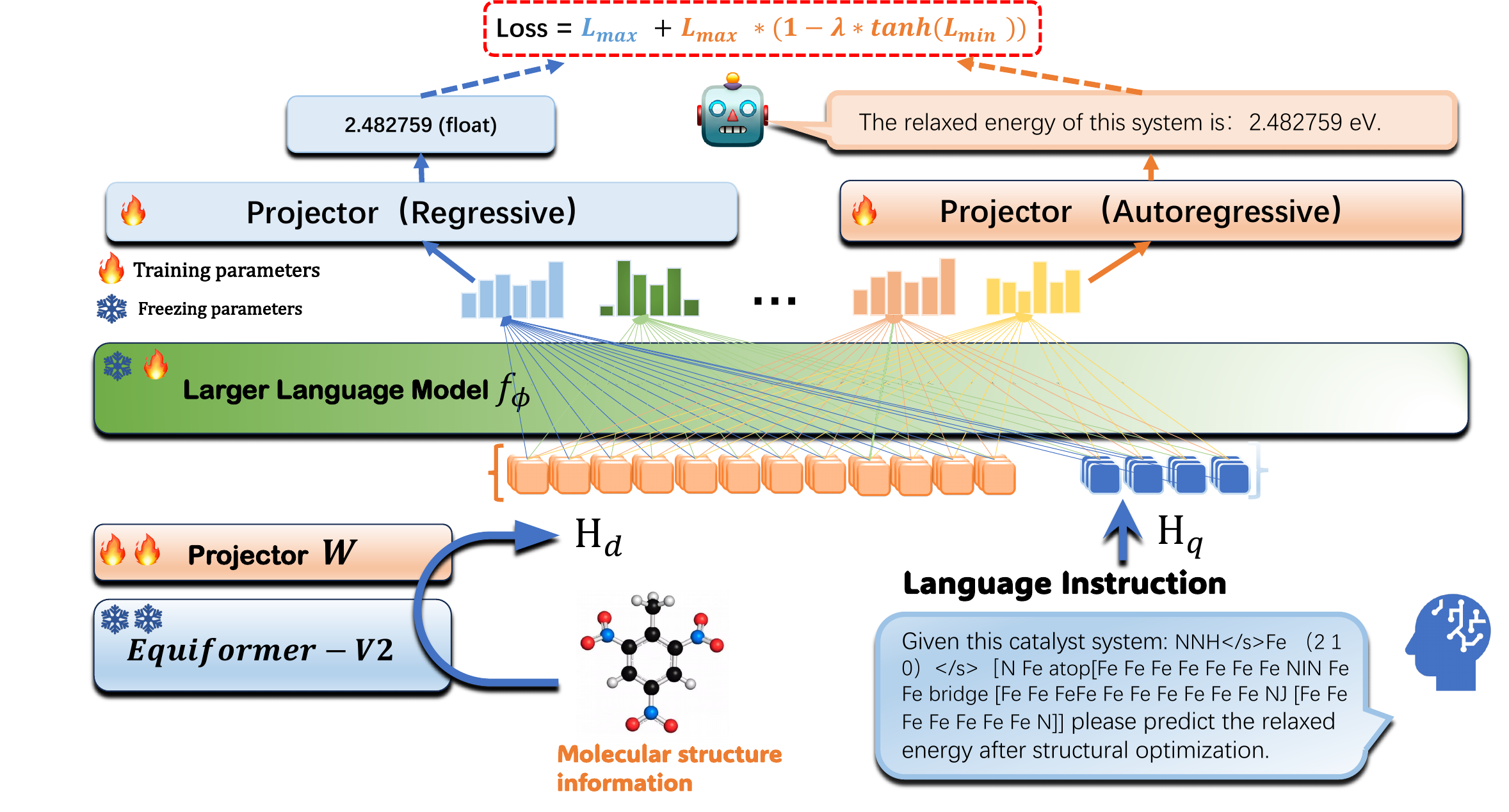}
\caption{
Schematic illustration of the QE-Catalytic framework. EquiformerV2 extracts geometric embeddings from 3D atomistic structures under periodic boundary conditions, while the structured text is encoded using a CatBERTa-style three-part string template. The two streams are first aligned in the latent space via contrastive learning, then fused within a multimodal LLM backbone, and finally jointly accomplish adsorption-energy prediction and structure generation through a numerical regression head and an autoregressive head.
}
\label{fig-1}
\end{figure*}
Existing approaches to adsorption-energy prediction can be broadly categorized into three paradigms. The first comprises graph-based equivariant GNNs exemplified by EquiformerV2, which take 3D atomistic graphs as input and directly regress system energies. These models excel at three-dimensional geometric modeling, but have difficulty integrating textual information. The second paradigm includes text-driven methods represented by CatBERTa, which are trained and inferred purely from structured strings. Such methods naturally absorb human-readable system information, yet lose explicit modeling of fine-grained 3D geometric details. The third paradigm consists of graph-assisted text methods such as GAP-CatBERTa, which align text embeddings using a geometric encoder during pretraining; however, at inference time, they remain essentially text-only models, unable to fully exploit available 3D structures and unable to completely resolve the ambiguity that ``the same text corresponds to multiple configurations.''
In this work, we propose QE-Catalytic, a multimodal large language model for catalytic materials property prediction. During inference, QE-Catalytic takes \emph{both} the 3D molecular structure and its corresponding textual description as inputs, thereby explicitly combining the geometric and semantic modalities, while still functioning properly when only the text modality is available. Specifically, as shown in \cref{fig-1}, QE-Catalytic employs EquiformerV2 as a 3D molecular feature extractor to encode atomistic structures under periodic boundary conditions; meanwhile, the structured textual representation of the catalytic system is provided to the multimodal LLM as part of the prompt. The geometric embedding is aligned with the text embedding via a linear projection, and the two are then fused through cross-modal interactions within a unified Transformer backbone. On the output side, we adopt a dual-head design\cite{dual1, dual2}: a numerical regression head that directly regresses relaxed adsorption energy, and an autoregressive head that generates energy-related text or structure tokens.

Training follows a three-stage pipeline. In Stage~1, we freeze the LLM parameters and train only EquiformerV2 and the projection layer, aligning geometric and textual embeddings via cross-modal contrastive learning to endow the model with preliminary predictive capability. In Stage~2, we unfreeze all parameters and conduct joint pretraining on a larger-scale multimodal dataset to further strengthen complementary modeling between graph and text. In Stage~3, we freeze EquiformerV2 and the projection layer and perform instruction fine-tuning only on the LLM, enabling QE-Catalytic to execute property-prediction and inverse-design instructions in natural language. With this design, the model can fully exploit geometric details when 3D coordinates are available, and degrades to a strong text-only predictor when graph is missing. Moreover, as shown in \cref{fig-2}, even when the textual information is incomplete, the model can generate a candidate CIF structure from the available text; the generated CIF can then be used to complement missing textual information of the catalytic system, thereby enabling property prediction.

\subsection{EquiformerV2 as the geometric encoder}

Equiformer is a class of $SE(3)$/$E(3)$-equivariant graph neural networks that combines equivariant inductive bias with the dynamic modeling capability of Transformers. The central idea is to replace the scalar operators in conventional Transformers with equivariant tensor operations on 3D atomistic graphs, and to introduce an equivariant graph-attention mechanism, thereby flexibly capturing local-environment information while preserving rotation/translation equivariance. EquiformerV2 further introduces multiple improvements: it replaces SO(3) convolution with eSCN convolution, and incorporates attention re-normalization, separable $S^2$ activations, and separable layer normalization. These designs substantially reduce computational cost under higher-order representations and achieve leading performance for energy and force prediction on the S2EF and IS2RE tasks of OC20.

In QE-Catalytic training, we adopt an EquiformerV2 model pretrained on the OC20 dataset as the geometric encoder for 3D molecular graphs, and extract graph embeddings after the final layer normalization and before the energy/force prediction heads. The model treats each atom as a node; each node corresponds to a two-dimensional embedding tensor, and the entire system is thus represented as a three-dimensional tensor. The size of this system tensor depends on the number of atoms, the number of spherical-harmonic channels, and the maximum spherical-harmonic order. Our embedding-extraction procedure is as follows: we first flatten each atom's 2D embedding into a 1D vector, and then apply max pooling across all atom vectors to obtain a single system-level embedding. To align with textual features, we project the embedding through a linear mapping head to the same dimensionality as the text embedding, which is used for subsequent graph--text contrastive learning and multimodal fusion. This design preserves EquiformerV2's equivariant geometric perception while enabling geometric features to participate in cross-modal attention within the LLM latent space as unified vectors.
\subsection{Max--Min tanh-gated multitask loss (MMTG-Loss)}
\label{subsec:mmtg_loss}

In this work, we jointly optimize two types of objectives: a regression loss $L_{\mathrm{MAE}}$ for continuous-property prediction (e.g., the MAE of adsorption energy), and a cross-entropy loss $L_{\mathrm{CE}}$ for discrete labeling or generation tasks. A conventional approach adopts a linear weighting form
\begin{equation}
    \label{eq:plain_loss}
    \mathcal{L}_{\mathrm{plain}}
    = \lambda\,L_{\mathrm{MAE}} + L_{\mathrm{CE}},
\end{equation}
where $\lambda>0$ is a manually specified weight. However, when the numerical scales of the two losses differ, or when the two losses converge at different rates across training phases, Eq.~\eqref{eq:plain_loss} often leads to the problem that \emph{one subtask dominates the gradients for an extended period, causing insufficient learning for the other subtask}, and the performance becomes highly sensitive to the choice of $\lambda$.

To address this issue, we propose a \emph{Max--Min tanh-gated multitask loss} (Max--Min Tanh-Gated Loss, MMTG-Loss). The key idea is to let the currently harder subtask dominate, while using the other subtask's loss as a bounded gating factor to smoothly modulate the overall loss. Specifically, we define
\begin{equation}
    L_{\max} = \max\!\bigl(L_{\mathrm{MAE}},\,L_{\mathrm{CE}}\bigr),\quad
    L_{\min} = \min\!\bigl(L_{\mathrm{MAE}},\,L_{\mathrm{CE}}\bigr)
\end{equation}
and define the total loss as
\begin{equation}
    \label{eq:mmtg_loss}
    \mathcal{L}_{\mathrm{MMTG}}
    = L_{\max} + L_{\max}\,\bigl(1 - \lambda\,\tanh(L_{\min})\bigr),
\end{equation}
where $\lambda\in(0,1]$ is a hyperparameter controlling the gating strength. Eq.~\eqref{eq:mmtg_loss} can be equivalently written as
\begin{equation}
    \mathcal{L}_{\mathrm{MMTG}}
    = L_{\max}\,\Bigl(2 - \lambda\,\tanh(L_{\min})\Bigr),
\end{equation}
which shows that $\mathcal{L}_{\mathrm{MMTG}}$ is always proportional to $L_{\max}$ and is modulated by $L_{\min}$ through the bounded function $\tanh(\cdot)$.

Compared with the linearly weighted loss $\mathcal{L}_{\mathrm{plain}}$, MMTG-Loss has several notable advantages:

\paragraph{(1) Automatically focusing on the currently hardest subtask.}
By explicitly introducing $L_{\max}=\max(L_{\mathrm{MAE}},L_{\mathrm{CE}})$, the overall loss is always dominated by the numerically larger term. Whether in early or late training, whenever one sub-loss is substantially larger, the corresponding subtask naturally takes gradient dominance, avoiding the ``imbalance'' caused by scale mismatch or improper weight selection in linear combinations.

\paragraph{(2) Smooth and bounded gating via $\tanh(L_{\min})$.}
The subtask corresponding to $L_{\min}$ is relatively ``easier.'' We map it to a bounded interval via $\tanh(L_{\min})\in(0,1)$ and construct the gating factor $1-\lambda\tanh(L_{\min})$. When both losses are large, $\tanh(L_{\min})\approx 1$ and the gating factor is approximately constant, so the model mainly focuses on reducing $L_{\max}$. As training progresses and $L_{\min}$ decreases, $\tanh(L_{\min})$ becomes smaller and the gating factor increases, gradually relaxing the penalty on $L_{\max}$ and yielding an adaptive ``prioritize stabilization first, refine later'' optimization strategy.

\paragraph{(3) Greater robustness to loss scales and reduced hyperparameter sensitivity.}
\begin{wrapfigure}{r}{0.49\textwidth}
  \centering
  \vspace{-20pt}
  \includegraphics[width=\linewidth]{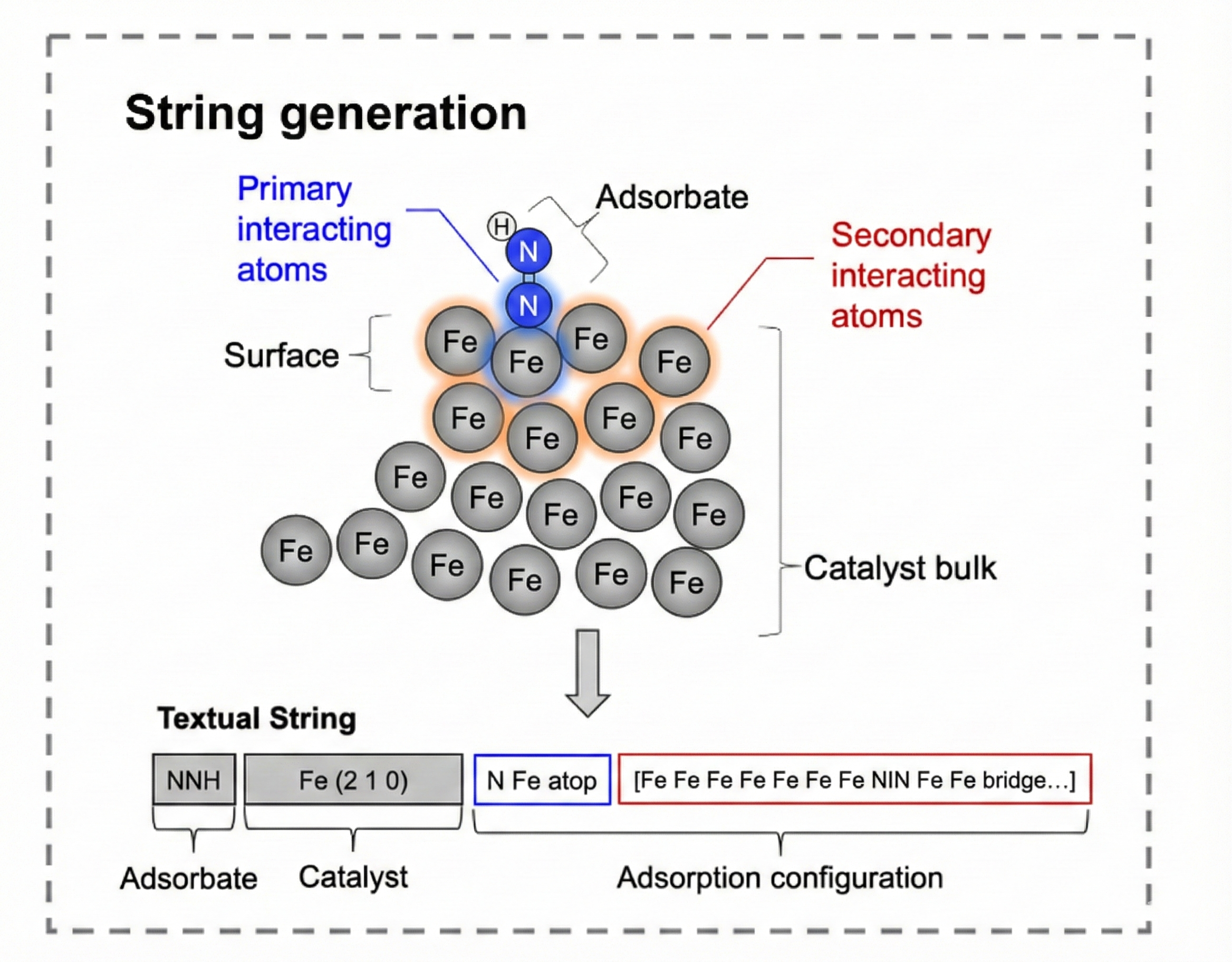}
  \caption{ This figure shows a process of converting a 3D structure into computer readable text}
  \label{fig_ablation}
\end{wrapfigure}
In conventional linear combinations, if the numerical scales of $L_{\mathrm{MAE}}$ and $L_{\mathrm{CE}}$ differ significantly, $\lambda$ requires careful tuning; even changing measurement units (e.g., from eV to meV) can necessitate re-searching. In contrast, MMTG-Loss uses $L_{\max}$ as the reference scale, and $L_{\min}$ only induces relative modulation through the bounded nonlinearity $\tanh(\cdot)$. Consequently, it is less sensitive to absolute loss scales and small perturbations of $\lambda$, yielding improved stability and transferability in practice.
\begin{figure*}[htp]
\centering
\includegraphics[width=136mm]{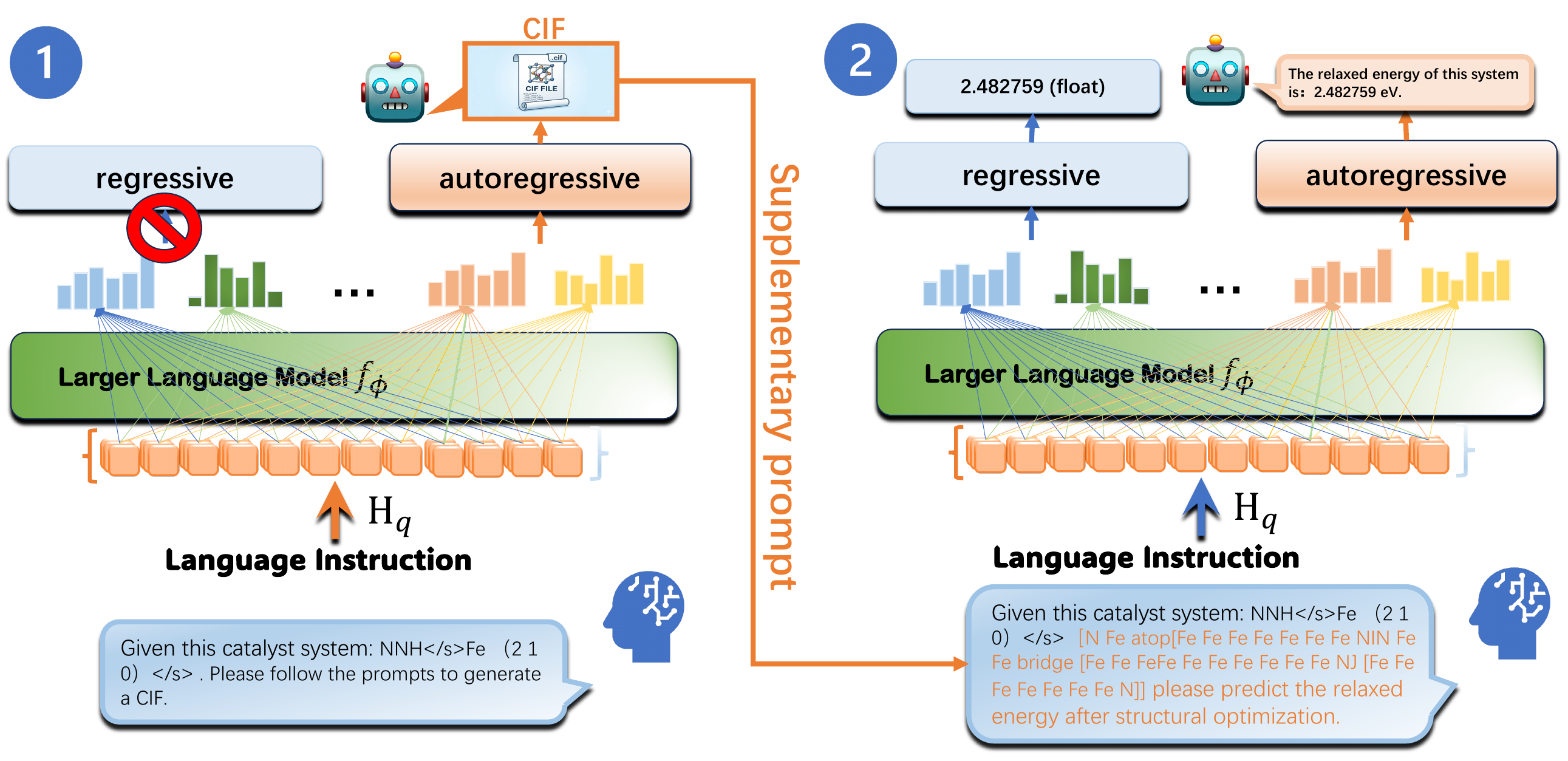}
\hspace{-9pt}
\caption{
Inverse-design workflow of QE-Catalytic when explicit 3D structures are unavailable. Given the first two text segments describing the adsorbate and the catalyst surface, the LLM autoregressively generates an indicative CIF. We then extract the local environment closest to the adsorbing atoms from the generated CIF and convert it back into a three-part configuration string for energy prediction or further screening.
}
\label{fig-2}
\end{figure*}
\paragraph{(4) Training dynamics better aligned with multitask-learning needs.}
From a gradient perspective, $\mathcal{L}_{\mathrm{MMTG}}$ is monotonically increasing in both sub-losses, while its effective weighting adapts across training stages. Early in training, when both losses are large, MMTG-Loss primarily suppresses the larger one; when a subtask is learned well (its loss decreases substantially), its influence is gradually gated down through $L_{\min}$ and $\tanh(L_{\min})$, allowing optimization to continue focusing on objectives that have not sufficiently converged. This property is conducive to more balanced convergence quality in multitask settings.
In summary, the Max--Min tanh-gated multitask loss $\mathcal{L}_{\mathrm{MMTG}}$ introduces a difficulty-aware adaptive weighting mechanism for multi-objective optimization while remaining simple to implement, and offers clear advantages over the conventional linearly weighted loss $\mathcal{L}_{\mathrm{plain}}$ in terms of stability, robustness, and multitask balance.
\subsection{Data format and structure-to-string conversion}

The text input in this study strictly follows the three-part string format proposed in the original CatBERTa paper. We convert relaxed structures from the OC20 and OC20-Dense datasets into text strings, as illustrated in the of \cref{fig_ablation}. The text input is organized into three segments: the adsorbate, the catalyst surface, and the adsorption configuration.

The \emph{adsorbate} segment contains only its element symbols. The \emph{catalyst surface} segment consolidates the overall catalyst composition and its Miller indices; both are obtained from the existing metadata of the OC20 dataset. The third segment, the \emph{adsorption configuration}, is expressed by identifying the primary and secondary interacting atoms, a strategy validated in prior work to be effective for energy prediction.

In implementation, we use the Pymatgen library to determine interaction factors. We first construct atomic connectivity based on a preset cutoff radius, which is taken as the covalent radius of each atom; we then identify atoms connected to the adsorbate atoms and the topmost-layer surface atoms. Atoms directly connected to the adsorbate atoms are classified as primary interacting atoms, while neighboring surface atoms around these primary interacting atoms are classified as secondary interacting atoms. Finally, we concatenate ``adsorbate element symbols--catalyst composition and Miller indices--primary/secondary neighbor configuration'' into a structured string, which serves as the standard input format for the QE-Catalytic text channel.

\subsection{Inverse-design data and indicative CIF generation}
Based on the above three-part data format, we further consider how to use QE-Catalytic for energy prediction and inverse design when explicit adsorbate--catalyst structures are unavailable. In this setting, we provide only the first two segments of the three-part text (the first segment is the chemical symbols of the adsorbate, and the second segment is the catalyst element symbols and the Miller indices of the exposed facet), and then let the large language model autoregressively generate the corresponding CIF file, as shown in Fig.~\ref{fig-2}. For the LLM, this is a conditional text-generation task: conditioned on ``adsorbate + catalyst + Miller indices,'' it generates a crystal-structure string with a plausible coordination pattern.

The generated CIF structure is not necessarily an atomically accurate relaxed configuration; rather, it is what we refer to as an ``indicative'' structure. That is, it is chemically plausible in terms of elemental composition and local neighborhood organization, but is not intended to be used directly as input to the GNN. When converting the LLM-generated CIF back into a configuration string, we adopt a more permissive and simplified procedure. We first identify only the adsorbate atoms closest to the surface; next, we collect the primary neighbor atoms surrounding the adsorbate atoms; then, from these primary neighbors, we collect secondary neighbor atoms. In this process, the cutoff radius is set to four times the covalent radius, i.e., atoms within this range are treated as neighbors. This strategy reflects the \emph{indicative} nature of generated CIFs: even when graph is imprecise, it preserves neighbor information that contributes to the local chemical environment, such that the resulting configuration string can still provide useful energy-prediction cues within QE-Catalytic.

\section{Training pipeline}

\subsection{Overall design}

QE-Catalytic is trained using a three-stage pipeline that follows the organizational form of Qwen2.5-VL while being customized to the multimodal characteristics of catalytic systems (text + 3D atomistic graphs). The overall objective is to enable the large language model, while preserving $SE(3)$/$E(3)$-equivariant geometric information, to: (i) jointly represent and leverage complementary information from text and 3D graphs for energy regression; (ii) perform robust inference when a sub-modality is missing; and (iii) execute bidirectional, instruction-following generation and inverse design between ``energy'' and ``structure strings/CIF.''

Notationally, we denote the model trained multimodally but using only text inputs at inference as \emph{QE-Catalytic*}, and the full model that uses both graph and text inputs at inference as \emph{QE-Catalytic}.

\subsection{Stage 1: graph--text alignment pretraining}

In the initial pretraining stage, QE-Catalytic primarily trains the 3D structural feature encoder (the EquiformerV2 branch) and the fully connected projection layer, while keeping the LLM parameters frozen. Using approximately 100k paired 3D-structure--configuration-text samples, the model aligns the 3D molecular features extracted by EquiformerV2 with the corresponding text embeddings via cross-modal contrastive learning, thereby establishing graph--text consistency in the latent space. The goal of this stage is to ensure that the model can stably capture key geometric information from 3D structures and establish one-to-one correspondence with textual descriptions, laying the foundation for subsequent joint pretraining and downstream tasks.

\subsection{Stage 2: multimodal joint pretraining}

In Stage~2, we unfreeze all parameters (including the LLM) based on the Stage~1 initialization, and scale up to a larger pretraining dataset. Compared with Stage~1, this stage introduces an additional $\sim$240k multimodal samples, covering a broader range of adsorbate--catalyst combinations and configuration diversity. Relative to the subset used in Stage~1, the Stage~2 dataset includes and extends earlier samples, thereby improving the model's ability to capture more complex geometric patterns and semantic conditions while maintaining graph--text alignment.

The training objective in this stage is to fully exploit complex interactions between 3D molecular structures and textual information within a unified multimodal backbone, such that the model can not only predict relaxed energies accurately when geometric information is available, but also fall back to reasonable performance when some modalities are missing. Through joint pretraining across diverse adsorption systems, QE-Catalytic acquires a more comprehensive representation capacity for the intricate relationships between 3D structures and text.

\subsection{Stage 3: instruction fine-tuning}

In the instruction fine-tuning stage, QE-Catalytic freezes the parameters of EquiformerV2 and the projection layer, and fine-tunes only the LLM component. We construct an instruction dataset of approximately 25k instances, including multimodal dialogue data and text-only dialogue data, covering multiple instruction types in a question--answer format, such as energy prediction and inverse generation. This multi-faceted data construction enables the model to learn to understand and execute natural-language instructions under different modality combinations (graph+text, text-only, etc.), thereby exhibiting stronger adaptability and robustness in real-world scenarios with missing modalities and diverse conditions.

\section{Code availability}
After the paper is accepted, we will open-source the source code
\section{Acknowledgments}

\paragraph{Funding:}
This work was supported in part by the National Natural Science Foundation of China under Grant 92370117, in part by CAS Project for Young Scientists in Basic Research under Grant YSBR-090, and in part by the Key Research Program of the Chinese Academy of Sciences under Grant XDPB22.

\paragraph{Competing interests:}
All authors of the article have no competing interests.

\bibliography{sn-bibliography}

\appendix

\end{document}